\documentclass[letterpaper, 10 pt, conference]{ieeeconf}  

\IEEEoverridecommandlockouts                              

\makeatletter
\let\NAT@parse\undefined
\makeatother
\usepackage[compress, numbers]{natbib}

\usepackage{amsmath,amsfonts,bm,amssymb}
\usepackage{xspace}
\usepackage{algorithm}
\usepackage{algpseudocode}

\usepackage{booktabs}

\newcommand{\alg}{\textsc{Sim-FSVGD}\xspace}
\newcommand{\sysid}{\textsc{SysID}\xspace}
\newcommand{\greybox}{\textsc{GreyBox}\xspace}

\def\eqref#1{equation~\ref{#1}}

\def\1{\bm{1}}

\newcommand{\norm}[1]{\left\lVert#1\right\rVert}

\def\rvh{{\mathbf{h}}}

\def\vf{{\bm{f}}}
\def\vg{{\bm{g}}}

\def\vs{{\bm{s}}}

\def\vu{{\bm{u}}}

\DeclareMathAlphabet{\mathsfit}{\encodingdefault}{\sfdefault}{m}{sl}
\SetMathAlphabet{\mathsfit}{bold}{\encodingdefault}{\sfdefault}{bx}{n}

\newcommand{\E}{\mathbb{E}}

\newcommand{\R}{\mathbb{R}}

\newcommand{\bX}{\mathbf{X}}

\newcommand{\bx}{\mathbf{x}}
\newcommand{\bs}{\mathbf{s}}

\newcommand{\bh}{\mathbf{h}}

\newcommand{\bu}{\mathbf{u}}

\newcommand{\bK}{\mathbf{K}}

\newcommand{\calD}{\mathcal{D}}

\newcommand{\calN}{\mathcal{N}}

\newcommand{\calX}{\mathcal{X}}
\newcommand{\calY}{\mathcal{Y}}

\usepackage{hyperref}
\usepackage{cleveref}
\usepackage{url}
\usepackage{adjustbox}
\usepackage{graphicx, caption}
\usepackage{caption}
\usepackage{subcaption}
\usepackage{times}
\usepackage{tikz}
\usepackage{lineno}
\usepackage{color}
\usepackage{hyperref}   
\usepackage{siunitx}

\title{\LARGE \bf
Bridging the Sim-to-Real Gap with Bayesian Inference
}

\author{Jonas Rothfuss$^{*, 1}$, Bhavya Sukhija$^{*, 1}$, Lenart Treven$^{*, 1}$, \\ Florian Dörfler$^{2}$, Stelian Coros$^{1}$, Andreas Krause$^{1}$
\thanks{*Equal contribution listed in alphabetical order}
\thanks{$^{1}$Department of Computer Science, ETH Zürich Switzerland
        {\tt\small email: firstname.lastname@inf.ethz.ch}}%
\thanks{$^{2}$Department of Electrical Engineering, ETH Zürich Switzerland
        {\tt\small email: dorfler@ethz.ch}}%
}

\begin{document}

\maketitle
\thispagestyle{empty}
\pagestyle{empty}

\maketitle

\begin{abstract}
\looseness=-1
We present \alg for learning robot dynamics from data. As opposed to traditional methods, \alg leverages
low-fidelity physical priors, e.g., in the form of simulators, to regularize the training of neural network models.
While learning accurate dynamics already in the low data regime, \alg scales and excels also when more data is available. We empirically show that learning with implicit physical priors results in accurate mean model estimation as well as precise uncertainty quantification. We demonstrate the effectiveness of \alg in bridging the sim-to-real gap on 
a high-performance RC racecar system.
Using model-based RL, we demonstrate a highly dynamic parking maneuver with drifting, using less than half the data compared to the state of the art.
\end{abstract}


\section{Introduction}
For many decades,
physical equations of motion have been leveraged to perform highly dynamic and complex tasks in robotics. 
However, recent advances have shown a significant gap between the physical models and the real world, also known as the `sim-to-real' gap. 
The gap is primarily because domain-specific models/simulators often tend to neglect complex real-world phenomena (e.g., aerodynamics, system latency, elastic deformation, etc.~\cite{widmer2023tuning, hwangbo}) 

Hence, in recent years, data-driven approaches, such as reinforcement learning (RL), have been widely applied in robotics~\cite{singh2022reinforcement}.
Particularly, model-based RL methods are often used for robot learning directly on hardware due to their improved sample-efficiency~\cite{deisenrothPILCOModelBasedDataEfficient2011, polydoros2017survey, wu2023daydreamer, bhardwaj2023data}.
In model-based deep RL, the dynamics of the real-world system are learned directly from data using
neural networks (NNs). However, these methods often ignore the abundance of knowledge available in physical simulators and instead directly fit a model from scratch. We believe that this approach is a critical source of data inefficiency. 

In this paper, we discuss how we can harness the benefits of both worlds, i.e., having the flexibility of NNs while retaining the domain knowledge embedded in simulators, encapsulating models of various degrees of fidelity.
Moreover, we leverage recent advances in Bayesian inference in functional space~\cite{wang2019function, sun2019functional} to impose a functional prior derived from the simulation/physical model. The prior regularizes the network to behave similarly to the simulation in a low-data regime and fit the real-world dynamics when data is available. Empirically, this results in a substantial sample-efficiency gain. 

\begin{figure}[t]
    \centering
    \includegraphics[width=\linewidth]{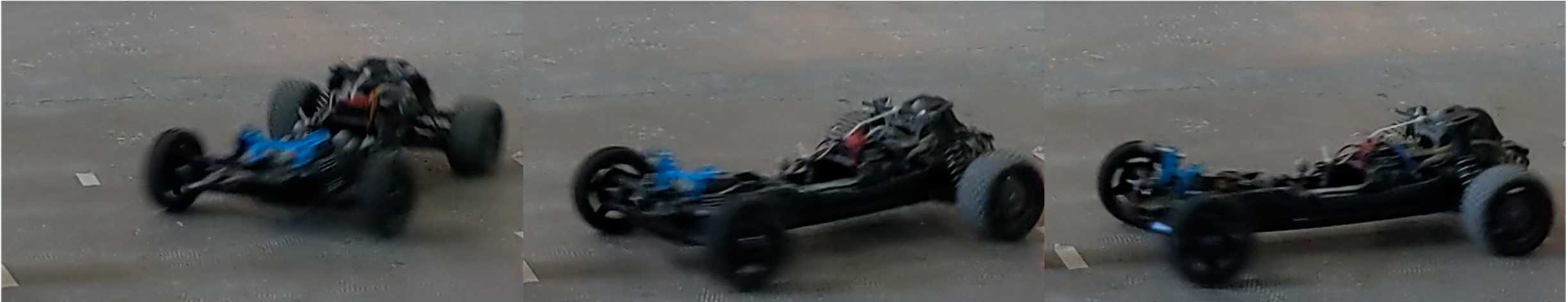}
    \caption{Dynamic RC car from our experiments.}
    \label{fig:teaser_rccar}
    \vspace{-0.5cm}
\end{figure}

In summary, our main contributions are; (\emph{i}) we propose \alg, a simple and tractable update rule for training Baysian Neural Networks (BNNs) that incorporates knowledge for available simulators, (\emph{ii}) we show that \alg has significant gains in sample-efficiency on learning real-world dynamics with NNs over several other widely applied methods, and (\emph{iii}) we combine \alg with a model-based RL algorithm and show that it results in considerably faster convergence on a highly dynamic RC Car (cf.,~\cref{fig:teaser_rccar}).

\section{Related Work. }
\subsubsection{Bridging the sim-to-real gap}
Methods such as domain randomization, meta-RL, and system identification are typically used to bridge the sim-to-real gap in robotics and RL (cf.,~\cite{zhao2020sim}). Domain randomization based approaches assume that the real world is captured well within the randomized simulator. In many cases, such as the ones we consider in this work, this is not the case. Meta-RL methods~\cite{duan2016rl, finn2017model, finn2018probabilistic, bhardwaj2023data} also require that the meta-training and test tasks are i.i.d. samples from a task distribution, and thereby representative. Furthermore, they generally require a plethora of data for meta-training, which is typically also computationally expensive. Instead, \alg does not require any meta training and solely leverages query access to low-fidelity simulators 
which are based on first-principles physics models and are often available in robotics. 
Moreover, we leverage solely the real world data and operate in the data-scarce regime. To this end, we consider model-based learning approaches due to their sample efficiency.
Typical model-based approaches, perform system identification~\cite{ljungSystemID, li2013terradynamics, moeckel2013gait, tan2016simulation, zhu2017model, gofetch} to fit the simulation parameters on real-world data. However, if the simulation model cannot capture the real-world dynamics well, this results in suboptimal performance (cf., ~\cref{sec:experiments}). 
Most closely related to our work are approaches such as~\cite{hwangbo, pastor2013learning,  ha2015reducing, HUANG20178969, schperberg2023real}. They use the simulator model and fit an additional model, e.g., NN or Gaussian process, on top to close the sim-to-real gap. We refer to this approach as \textsc{GreyBox}. In \cref{sec:experiments} we show that our method results in better uncertainty estimates than the \textsc{GreyBox} approach. 

\subsubsection{Neural Networks for System Identification}
\looseness=-1
Neural networks are commonly used for learning robot dynamics from data~\cite{hwangbo, narendraNNSysID, SJOBERG1994359, nagabandi2018, bernSoftRobotControl2020, sukhija2023gradient} due to their high expressiveness. 
However, during control, planning with NN models is challenging because most planners exploit (``overfit'' to) inaccuracies of the learned model. This leads to suboptimal performance on the real system~\cite{chua2018deep}.
BNNs do not suffer from the same pitfall~\cite{chua2018deep}. As a result, recently, BNNs have been widely used for dynamics learning~\cite{chua2018deep, curi2020efficient, sekar2020planning, treven2024ocorl,
sukhija2024optimistic,rothfuss2023hallucinated}. 

\subsubsection{Bayesian Neural Networks}
Unlike learning a single NN, BNNs maintain a distribution over the NN parameters. Moreover, using a known prior distribution over the NN parameters, given the training data, BNNs infer the posterior distribution. Exact Bayesian inference is computationally intractable for NNs, and approximations~\cite{mackay1992bayesian, neal2012bayesian, graves2011practical,blundell2015weight, gal2016dropout, zellers2018swag, goan2020bayesian} are generally used.
In this work, we focus on particle-based BNNs due to their computational tractability and reliable uncertainty estimates ~\cite{wang2019function, lakshminarayanan2017simple, Liu2016} (cf.,~\cite{d2021repulsive} or \cref{sec:background_fbnns} for more details). 
Nearly all BNNs impose a generic standard normal prior on the NN parameters. On the contrary, we leverage existing (low-fidelity) simulators to impose an informed prior. This boosts sample efficiency when learning robot dynamics from data and also enables bridging the `sim-to-real' gap (cf.,~\cref{sec:experiments}).

\section{Background} 
\label{sec:background_fbnns}


\subsection{Learning Dynamics Models with BNNs}
The main goal of this work is to efficiently learn a robot dynamics model from data, that is, given a dataset of state-action\footnote{Analogous to model-based learning literature, we focus on the fully observable setting.} pairs and the next state $(\vs_t, \vu_t, \vs_{t+1})$, learn a model to predict the next state. To this end, we focus on a general regression problem with a dataset $\mathcal{D}=(\mathbf{X}^{\mathcal{D}},\mathbf{y}^{\mathcal{D}})$, comprised of $m$ noisy evaluations $y_{j}=f\left(\bx_{j}\right) + \epsilon_j$ of an unknown function $f: \calX \mapsto \calY$. Here, the training inputs are denoted by $\mathbf{X}^{\mathcal{D}} = \left\{ \bx_j \right\}_{j=1}^m$ and corresponding function values by $\mathbf{y}^{\mathcal{D}} = \left\{ y_j \right\}_{j=1}^m$. The noise is assumed to be i.i.d.~and Gaussian.
To fit a regressor on the data, we employ a NN $h_\theta: \calX \rightarrow \calY$ with weights $\theta \in \Theta$. The conditional predictive distribution for the noisy observations can be defined as $p(y|\bx,\theta) = \calN(y|h_\theta(\bx), \sigma^2)$ using $h_\theta$, with $\sigma^2$ as the observation noise variance.

BNNs maintain a distribution over the NN parameters $\theta$. In particular, Bayes' theorem combines a prior distribution $p(\theta)$ with the empirical data into a posterior distribution $p(\theta|\mathbf{X}^{\mathcal{D}}, \mathbf{y}^{\mathcal{D}}) \propto p(\mathbf{y}^{\mathcal{D}}|\mathbf{X}^{\mathcal{D}},\theta) p(\theta)$. Here, $p(\mathbf{y}^{\mathcal{D}}|\mathbf{X}^{\mathcal{D}},\theta)$ is the likelihood of the data, given a NN hypothesis $\theta$. Under the i.i.d.~hypothesis, the likelihood factorizes as $p(\mathbf{y}^{\mathcal{D}}|\mathbf{X}^{\mathcal{D}},\theta) = \prod_{j=1}^m p(y_j|\bx_j,\theta)$.

To make predictions for an unseen test point $\bx^*$, we marginalize out the parameters $\theta$ in the posterior, i.e., the predictive distribution is calculated as 
\begin{align*}
    p(y^*|\bx^*,\mathbf{X}^{\mathcal{D}}, \mathbf{y}^{\mathcal{D}}) &= \int p(y^*|\bx^*,\theta) p(\theta|\mathbf{X}^{\mathcal{D}}, \mathbf{y}^{\mathcal{D}}) d\theta \\
    &= \E_{\theta}\left[ p(y^*|\bx^*,\theta) | \mathbf{X}^{\mathcal{D}}, \mathbf{y}^{\mathcal{D}} \right].
\end{align*}

\subsection{BNN Inference in the Function Space}
\looseness -1 Due to the high-dimensional parameter space $\Theta$ and the over-parameterized nature of the NN mapping $h_\theta(\mathbf{x})$, posterior inference for BNNs is challenging. Hence, we need to fall back on some form of {\em approximate inference} \cite{Blei2016, chen2018unified} to make BNN inference tractable. Moreover, the complex parametrization also makes choosing an appropriate prior distribution $p(\theta)$ very hard.

Aiming to alleviate these issues, an alternative approach views BNN inference in the function space, i.e., instead of imposing a prior in the parameter space $\Theta$, a prior in the 
space of regression functions $h: \calX \mapsto \calY$ is used. This results in the posterior $p(h |\mathbf{X}^{\mathcal{D}}, \mathbf{y}^{\mathcal{D}}) \propto p(\mathbf{y}^{\mathcal{D}} | \mathbf{X}^{\mathcal{D}},h) p(h)$ \cite{wang2019function, sun2019functional}. Here, $p(h)$ is a {\em stochastic process} prior, e.g., Gaussian process (GP)~\cite{rasmussen2003gaussian} with index set $\calX$, taking values in $\calY$. 

\looseness -1 Stochastic processes can be understood as infinite-dimensional random vectors. Hence, they are computationally intractable. However, given finite {\em measurement sets} $\bX:= [\bx_1, ..., \bx_k] \in \calX^k, k \in \mathbb{N}$, the stochastic process can be characterized by its marginal distributions of function values $\rho(\mathbf{h}^{\mathbf{X}}):= \rho(h(\bx_1), ... h(\bx_k))$ (cf. Kolmogorov Extension Theorem, \cite{stochastic_differential_equations}). This allows us to break down functional BNN inference into a more tractable form by re-phrasing it in terms of posterior marginals for measurement sets $\bX$:
$p(\rvh^\bX |\mathbf{X}, \mathbf{X}^\mathcal{D}, \mathbf{y}^{\mathcal{D}})  \propto  p(\mathbf{y}^{\mathcal{D}} | \rvh^{\bX^{\mathcal{D}}}) p(\rvh^{\bX})$. 

\looseness -1 This functional posterior can be tractably approximated for BNN inference \cite{wang2019function, sun2019functional}. In the following, we briefly describe how this can be done with functional Stein Variational Gradient Descent \cite[FSVGD,][]{wang2019function}:
The FSVGD method approximates the posterior as a set of $L$ NN parameter particles $\{ \theta_1, ..., \theta_L \}$. To improve the particle approximation, FSVGD iteratively re-samples measurement sets $\bX$ from a distribution $\nu$ supported on $\calX$, e.g.,~\textsc{Uniform}$(\calX)$. It then computes SVGD updates \cite{Liu2016} in the function space, and projects them back into the parameter space, to update the particle configuration.

To guide the particles towards areas of high posterior probability, FSVGD uses the functional posterior score, i.e., the gradient of the log-density
\begin{align}
    \nabla_{\rvh^{\bX}} &\ln p(\rvh^{\bX} |\mathbf{X},  \mathbf{X}^\mathcal{D},\mathbf{y}^{\mathcal{D}})  = \notag \\ &\nabla_{\rvh^{\bX}} \ln p(\mathbf{y}^{\mathcal{D}} | \rvh^{\bX^{\mathcal{D}}}) + \nabla_{\rvh^{\bX}} \ln p(\rvh^{\bX}) ~.
\end{align} In particular, for all $l = 1, ..., L~$ the particle updates are computed as
\begin{align} \label{eq:fsvgd_updates}
    &\theta^l \leftarrow \theta^l - \gamma J_l u_l, \quad \text{where }  J_l = \underbrace{\left( \nabla_{\theta^l} \rvh_{\theta^l}^\bX \right)^\top}_{\text{NN Jacobian}}  \\
    &u_l =  
    \bigg( \underbrace{\frac{1}{L} \sum_{i=1}^L \bK_{li} \nabla_{\rvh_{\theta^i}^\bX} \ln p(\rvh_{\theta^l}^\bX |\mathbf{X}, \mathbf{X}^\mathcal{D}, \mathbf{y}^{\mathcal{D}}) + \nabla_{\rvh_{\theta^l}^\bX} \bK_{li}}_{\text{SVGD update in the function space} \vspace{-4pt}} \bigg), \notag
\end{align}
where $\mathbf{K} = [k(\rvh_{\theta^l}^\bX, \rvh_{\theta^i}^\bX)]_{li}$ is the kernel matrix between the function values in the measurement points based on a kernel function $k(\cdot, \cdot)$.
As we can see in (\ref{eq:fsvgd_updates}), FSVGD only uses the prior scores and, in principle, does not require the density function of the prior marginals. This constitutes a key insight that we will later draw upon in our approach.




\section{\alg}
In this section, we present \alg. We start this section with an illustrative example to motivate our approach.
\begin{algorithm*}[th]
\caption*{\textbf{Estimating Functional Prior from Physical Prior}}\label{alg:test}
\textbf{Input:} Measurement distribution
$\nu$, Physical prior 
$\bm{g}$, Parameter distribution $p(\bm{\phi})$. 
\begin{algorithmic}[1]

\State Sample measurement set $\bX_M \sim \nu$
\State Sample parameters $\bm{\phi}_j$ for $j \in \{1, \dots, P\}$ from $p(\bm{\phi})$.
\State Calculate physical model output $\vg(\bX^M; \bm{\phi}_j)$ for $j \in \{1, \dots, P\}$.
\State Calculate empirical mean $\mathbf{\mu}({\bX_M})$ and variance $\mathbf{\Sigma}({\bX_M})$ over the $P$ samples.
\State Approximate Gaussian prior $p(\bh^{\bX_M}) \sim \mathcal{N}(\mathbf{\mu}({\bX_M}), \mathbf{\Sigma}({\bX_M}))$ 
\end{algorithmic}
\end{algorithm*}

\begin{algorithm}[th]
\caption{\strut \alg}\label{alg:sim_transfer_alg}
\hspace*{\algorithmicindent} \textbf{Input:} Measurement distribution
$\nu$, Sim prior \\
\hspace*{\algorithmicindent} \hspace{2.7em} $g$, Parameter distribution $p(\phi)$, GP \\ 
\hspace*{\algorithmicindent} \hspace{2.7em} $p(\tilde{\bh})$, Data $\calD$, BNN particles $\{\theta_i\}^{L}_{i=1}$
\begin{algorithmic}[1]

\State Sample measurement set $\bX \sim \nu$
\State Sample sim finctions $\{\bh^{\bX}_{i, j}\}^{d_y}_{i=0}$ for $j \in \{1, \dots, P\}$.
\State Calculate empirical mean $\{\mathbf{\mu}^{\bX}_i\}^{d_y}_{i=1}$ and variance $\{\mathbf{\Sigma}^{\bX}_i\}^{d_y}_{i=0}$ over the $P$ samples.
\State Approximate Gaussian prior for each $i$, i.e, $p(\bh^{\bX}_i) \sim \mathcal{N}(\mathbf{\mu}^{\bX}_i, \mathbf{\Sigma}^{\bX}_i)$ 
\State Update BNN particles with \cref{eq:fsvgd_updates}.
\end{algorithmic}
\end{algorithm}


\subsection{Illustrative Example: the Pendulum}
Let us consider a simplified pendulum model where the pendulum's state $\bs = [\theta, \dot{\theta}]$ is comprised of the angle $\theta$ and angular velocity $\dot{\theta}$. We can describe the change of state (i.e., transition function) by the ODE
\begin{align} \label{eq:pendulum_model}
    \ddot{\theta} = \frac{mgl \sin(\theta) + C_m u }{I} ~.
\end{align}
Here $m, l$ and $I$ are the mass, length, and moment of inertia, respectively. 
The motor at the rotational joint of the pendulum applies the torque $\tau = C_m u$ proportional to the control input $u$. 


When aiming to accurately predict how the real pendulum system behaves, there are two key sources of uncertainty/error. First, the exact parameters $\phi$ (here, $\phi=[m, l, C_m, I]$) of the domain-specific model are unknown. 
However, we can typically narrow down each parameter's value to a plausible range. This can be captured via a prior distribution $p(\phi)$ over the model parameters. The process of randomly sampling a parameter set $\phi \sim p(\phi)$ and then integrating/solving the ODE in \Cref{eq:pendulum_model} with the corresponding parameters gives random functions. This allows us to implicitly construct a stochastic process of functions that reflect our simplified pendulum model. We call this the {\em domain-model process}.

The second source of uncertainty is various physical phenomena such as aerodynamic drag, friction, and motor dynamics that are not captured in (\ref{eq:pendulum_model}). This results in a systematic sim-to-real gap. 
When constructing a prior $p(h)$ for training a BNN model on the real system, we also need to model the sim-to-real gap. 


\subsection{Main Algorithm}

Now we describe our approach, \alg, which combines the domain-model process with a sim-to-real prior to construct an informed functional prior $p(h)$ for learning the dynamics model. The final algorithm is presented in \Cref{alg:sim_transfer_alg}. 
First, we summarize how to construct the stochastic process prior.

In practice, we aim to model vector-valued functions, i.e., $h: \calX \mapsto \calY$ where $\calX \subseteq \R^{d_x}$ and $\calY \subseteq \R^{d_y}$. We factorize the prior over the output dimensions, i.e., $p(h) = \prod_{i=1}^{d_y} p(h_i)$. This allows us to treat each $h_i: \bX \mapsto \R$ as an independent scalar-valued function.

\subsubsection{Domain-model process} The first component of the implicit prior is the domain-specific (low-fidelity) simulation model $g(\bx, \phi)$ which takes as an input the state action pair $(\bs, \bu)$, i.e., $\bx = [\bs, \bu]$, as well as system parameters $\phi$. Crucially, we do not require an analytical expression of $g: \calX \mapsto \calY$. It is sufficient to have query access, i.e., the output vector of $g(\bx, \phi)$ can be the result of a numerical simulation (e.g., ODE solver or rigid body simulator). 

\subsubsection{Sim-to-real prior} The second component is the sim-to-real gap process. For that we employ a GP $p(\tilde{h}_{i})$ per output dimension $i=1, ..., d_y$ which aims to model the residual/gap between the best possible domain model $g_i(\bx, \phi^*)$ and the actual target function $f(\bx)$. We use a GP with zero mean and an isotropic kernel $k(\bx, \bx') = \nu^2 \rho(\norm{\bx-\bx'} / \ell)$ with variance $\nu^2$ and lengthscale $\ell$. 
Accordingly, the marginal distributions follow a multivariate normal distribution $p(\tilde{\bh}^{\bX}_{i}) = \calN(\tilde{\bh}^{\bX}_i | \mathbf{0}, \bK )$ with kernel matrix $\bK = [k(\bx_i, \bx_j)]_{i,j}$. The lengthscale $l$ and $\nu^2$ are hyperparameters that influence our belief on the sim-to-real gap.
 A small $\nu^2$ implies that the sim-to-real gap is with high probability not too big 
 and a large lengthscale conveys that deviations from the domain model are systematic rather than local. 

We combine both processes by independently sampling (conditional) random vectors from both processes and adding them, i.e,
\begin{align} \label{eq:combined_sampling}
\begin{split}
\bh_i^{\bX} = \hspace{2pt} & [g_i(\bx_1, \phi), ..., g_i(\bx_k, \phi)]^\top + \tilde{\bh}^{\bX}_i \\ & \text{with}~~ \phi \sim p(\phi) , ~  \tilde{\bh}_i^{\bX} \sim \calN(\tilde{\bh}_i^{\bX}| \mathbf{0}, \bK ).
\end{split}
\end{align}
\looseness -1
The resulting stochastic process prior $p(h)$ is defined implicitly through the marginal distributions that are implied by (\ref{eq:combined_sampling}). 

\subsubsection{Estimating the Stochastic Process Score}
To perform FSVGD updates as in (\ref{eq:fsvgd_updates}), we have to compute the score of our stochastic process prior, i.e., $\nabla_{\rvh^\bX} \ln p(\rvh^{\bX}) = \sum_i^{d_y} \nabla_{\rvh_i^\bX} \ln p(\rvh_i^{\bX})$. This is intractable for most domain-model processes and we need to estimate it from samples. 
To this end, we use a simple and efficient Gaussian approximation for the score of prior marginals.
In particular, we sample a measurement set $\bX$ from a known measurement distribution $\nu$, e.g., uniform distribution over the state-action space.  
For the measurement set $\bX$, we sample $P$ vectors of function values $\bh_{i,1}^{\bX}, ...\bh_{i,P}^{\bX} \sim p(\bh_i^{\bX})$ via (\ref{eq:combined_sampling}). Then, we compute their mean and covariance matrix, that is $\mathbf{\mu}^{\bX}_i$ and $\mathbf{\Sigma}^{\bX}_i$ and use the score of the corresponding multivariate Gaussian as an approximation, i.e., $p(\bh^{\bX}_i) \sim \mathcal{N}(\mathbf{\mu}^{\bX}_i, \mathbf{\Sigma}^{\bX}_i)$. 

\section{Experiments} \label{sec:experiments}
In this section, we present our experimental results where we evaluate our method, \alg, on  
learning the robot dynamics from data and online reinforcement learning.
\paragraph{Baselines} We consider three baselines; (\emph{i}) FSVGD~\cite{wang2019function}, (\emph{ii}) \sysid, and (\emph{iii}) \greybox. As discussed in 
\cref{sec:background_fbnns}, FSVGD is widely applied for Bayesian inference in deep learning. 
\sysid is a simple baseline, that uses the data from the real system to fit the parameters of the simulation model. 
Crucially, \sysid cannot capture the sim-to-real gap. To this end, we also consider \greybox which first fits the simulation model and then learns a BNN to fit the difference between the real data and the sim model, akin to~\cite{pastor2013learning, HUANG20178969, schperberg2023real}.

\subsection{One-Dimensional Experiment}
\begin{figure*}[ht!]
    \centering
    \includegraphics[width=\textwidth]{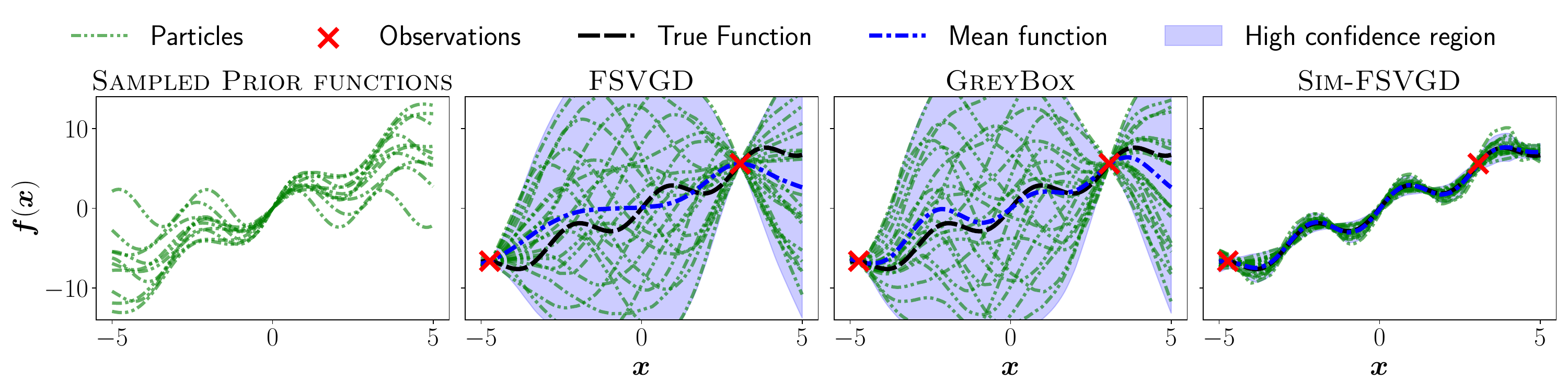}
    \caption{BNN posteriors trained on two data points for a one-dimensional sinusoidal function. \textsc{GreyBox} finds accurate mean, but has large uncertainty estimates. At the same time \alg obtains both accurate mean and uncertainty estimates.}
    \label{fig:1d_sinusoids}
\end{figure*}
We consider a one-dimensional sinusoidal function to visualize the difference between our method and the baselines in \cref{fig:1d_sinusoids}. Moreover, we train the BNNs with only two data points and plot their posteriors. From \cref{fig:1d_sinusoids} we conclude that FSVGD and \greybox result in very general posteriors that do not capture the sinusoidal nature of the true function. 
On the contrary, \alg can learn an accurate posterior with only two data points. 

\begin{figure*}[ht]
    \centering
    \includegraphics[width=\linewidth]{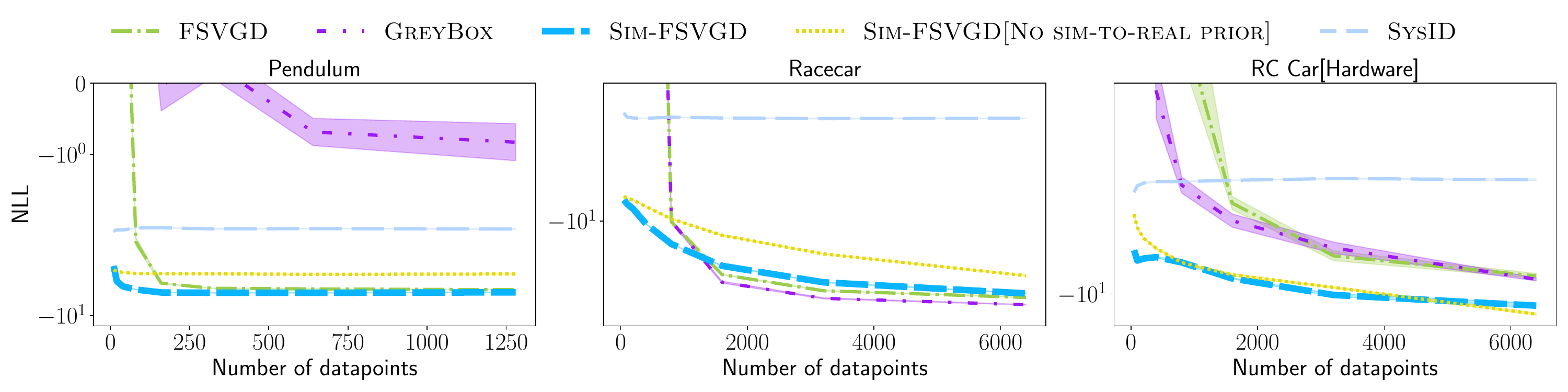}
  \caption{\looseness-1 We benchmark NLL of dynamics model on test data for Racecar on hardware and in simulations and Pendulum in simulation. BNNs with simulation priors achieve low NLL already in small data regimes and constantly outperform the standard BNNs when we increase the number of train data points.}
  \label{fig:regression_real_exp}
\end{figure*}
\subsection{Sim to Real Transfer}
\looseness -1
In this experiment, we investigate how \alg bridges the sim-to-real gap for dynamical systems. To this end, we consider a pendulum and racecar in simulation as well as a real-world RC car as our systems. We use an i.i.d.~dataset of sampled transitions from the real system to fit a model of the dynamics. As the evaluation metric, we consider the negative log-likelihood (NLL) of our model on a test dataset.
For the pendulum, we use the model from~\cref{eq:pendulum_model} as the simulation prior and we simulate the sim-to-real gap by incorporating aerodynamic drag, friction, and motor dynamics in the `real' dynamics. For the racecar, we use the dynamics model from~\cite{kabzan2020amz} which models tire dynamics with the Pacejka tire model~\cite{pacejka1992magic}. For the simulation prior, we use the kinematics bicycle model also from~\cite{kabzan2020amz}, which does not capture the tire dynamics. We use the same simulation prior for real RC car. The car consists of a high-torque motor, which allows 
us to perform dynamic maneuvers that involve loss of traction and drifting. These behaviors are neither captured by our simulation prior nor the dynamics model from~\cite{kabzan2020amz}.

The RC car is similar to the one in~\cite{sukhija2023gradient, bhardwaj2023data} and has a six-dimensional state (position, orientation, and velocities) and two-dimensional input (steering and throttle). We control the car at \SI{30}{\hertz}. and use the Optitrack for robotics motion capture system
\footnote{\href{https://optitrack.com/applications/robotics/}{https://optitrack.com/applications/robotics/}}
to estimate the state. 
The car has a significant delay (ca. \SI{80}{\milli \second})
between transmission and execution of the control signals on the car. 
Therefore, we include the last three actions $[a_{t-3}, a_{t-2}, a_{t-1}]$ in the current state $s_t$. The resulting state space is $12D$ and the action space is $2D$.

\looseness=-1
Our results are presented in \cref{fig:regression_real_exp}. From the figure, we conclude that \alg consistently outperforms all other methods in both case studies. Moreover, in the low-data regime, our method achieves orders of magnitude lower NLL, even though there is a considerable sim-to-real gap (c.f., \sysid). Furthermore, when more data is provided our method performs similarly to other baselines. Therefore, \alg relies less on the simulation prior when more data is available. This
empirically validates the consistency of our approach.
In \cref{fig:regression_real_exp}, we also study the effect of incorporating the additional sim-to-real GP prior in \alg. We conclude that including the additional GP leads to better performance and thus plays a crucial role, particularly for the pendulum and racecar. 



\subsection{Online RL}
\begin{figure*}[ht!]
    \centering
    \includegraphics[width=\textwidth]{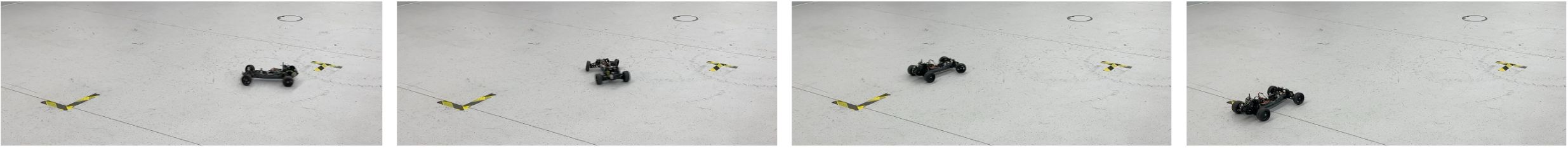}
    \caption{Desired reverse parking maneuver which involves rotating the car \SI{180}{\degree} and parking ca.~\SI{2}{\meter} away. }
    \label{fig:parking_demo}
\end{figure*}

\begin{figure}[ht!]
    \centering
   {\input{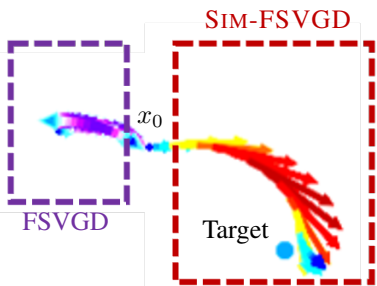}}
\vspace{-2em}
    \caption{\looseness-1 Realized trajectory for Racecar environment in simulation after episode 3. In violet, we have the trajectory of the model with FSVGD (no prior), and in red the trajectory of \alg using the low-fidelity prior. We observe that,
    \alg already learns to get close to the target in three episodes, whereas FSVGD (no prior) explores away from it.}
  \label{fig:realized_trajectory}
\end{figure}

\begin{figure*}[ht!]
   \begin{center}
\input{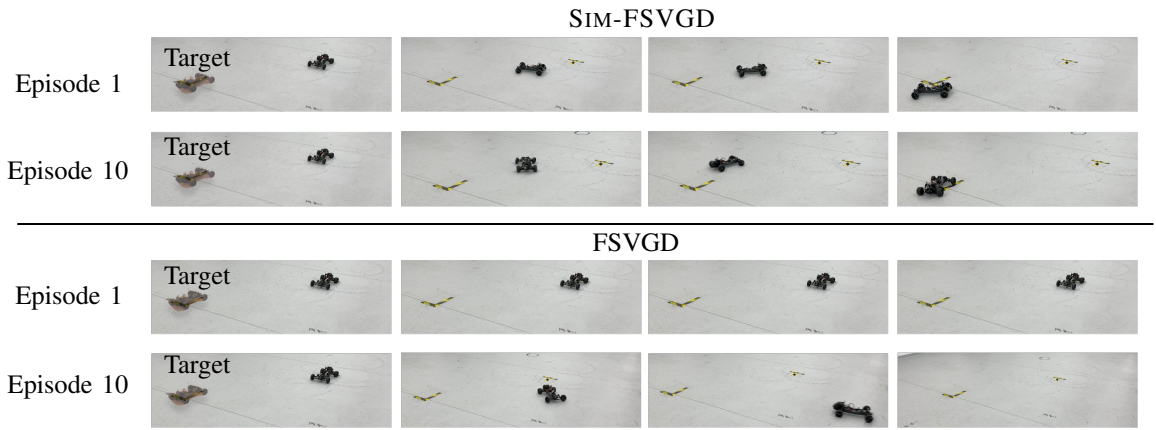}
\end{center}
\caption{Realized trajectories after episodes one and ten for \alg and FSVGD. \alg moves close to the target in episode one and parks nearly perfectly in episode ten. On the contrary, FSVGD fails to move at all in the first episode and explores away from the target in the tenth.}
\label{fig:hw_online_rl}
\vspace{-0.2cm}
\end{figure*}

\begin{figure}[ht!]
    \centering
    \includegraphics[width=\linewidth]{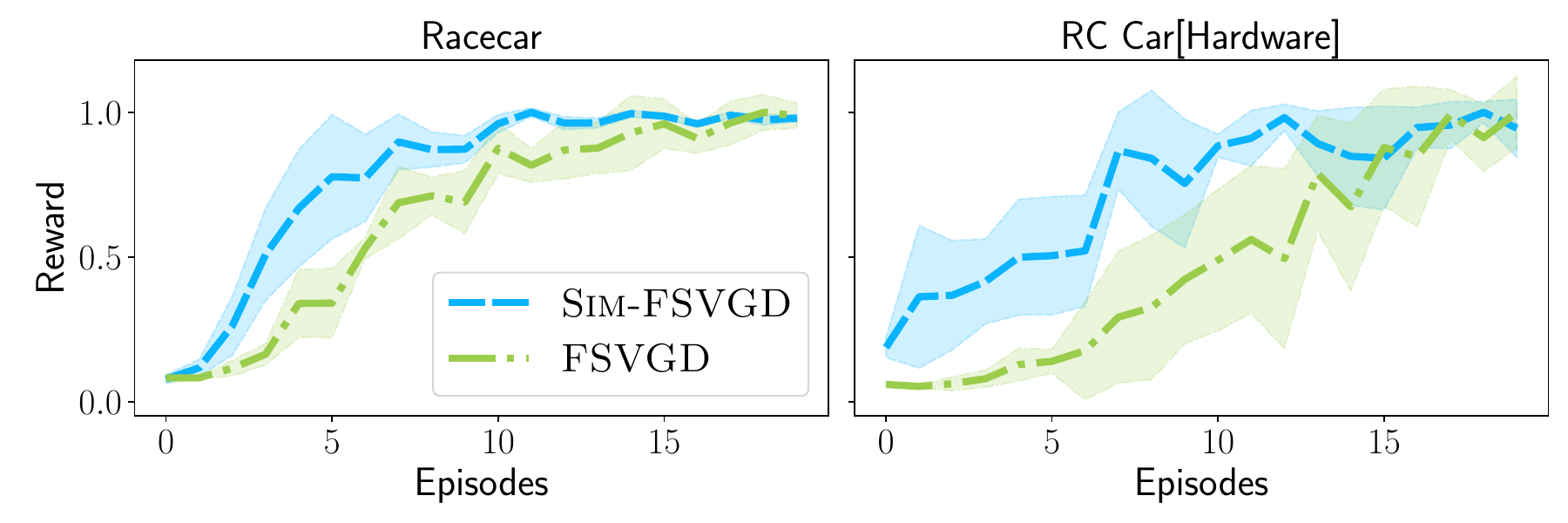}
    \caption{The simulation prior helps initially by exploring more directly towards the target. Both in simulation and on hardware the agent solves the task significantly faster and with less collected data.}
    \label{fig:online_rl}
    \vspace{-0.6cm}
\end{figure}

\looseness -1
We study the problem of model-based reinforcement learning for the racecar in simulation and the real RC car. In model-based RL, BNNs with good uncertainty estimates play a crucial role in exploration~\cite{chua2018deep, curi2020efficient, sekar2020planning, treven2024ocorl, sukhija2024optimistic}. Moreover, incorporating a simulation prior can lead to directed exploration and therefore faster convergence, i.e., better sample efficiency. To this end, in this experiment, we compare the performance of \alg and FSVGD in model-based RL. Moreover, we use our BNN to obtain a policy $\pi$, which we rollout on the real system to get further data for updating our model (see~\cref{alg:rl_sim_transfer}). We repeat this procedure for 20 episodes and consider a rollout length of $100$. For the policy $\pi$, we use the SAC algorithm~\cite{sac} which we train using our BNN similar to~\cite{janner2019trust}.
\begin{algorithm}[th]
\caption{\strut \alg for Model-Based RL}\label{alg:rl_sim_transfer}
\hspace*{\algorithmicindent} \textbf{Input:} Initial model and policy $\vf_\Theta$, $\pi_{\psi}$, $\calD := \emptyset$ 
\begin{algorithmic}[1]

\For{$\textnormal{episode}=1,2,\ldots$}
\State $\vf_{\Theta} \leftarrow \text{\alg}(\vf_\Theta, \calD)$  
\State $\pi_{\psi} \leftarrow \text{SAC}(\pi_{\psi}, \vf_{\Theta})$ 
    \State $(s_0, a_0, ..., a_{T-1}, s_T) \leftarrow \textsc{Rollout}(\pi_{\psi})$ 
    \State $\calD \leftarrow \calD \cup \{(s_t, a_t, s_{t+1})\}_{t=0}^{T-1}$  
\EndFor
\end{algorithmic}
\end{algorithm}
\looseness -1
The RL task we consider involves reverse parking the car ca. \SI{2}{\meter} away from the initial position $x_0$ by quickly rotating the car \SI{180}{\degree}.
(c.f.,~\cref{fig:parking_demo} or the video demonstration\footnote{
\href{https://crl.ethz.ch/videos/draft_iros_low_res.mp4}{https://crl.ethz.ch/videos/iros\_video}}). This typically leads to the car drifting, a behavior that is not captured by our simulation prior. 

\looseness -1 In \Cref{fig:online_rl} we report the learning curves for the racecar in simulation and the real RC car. We observe that in both cases, incorporating the simulation prior leads to considerably faster convergence (ca.~$2\times$ fewer episodes). Moreover, in \Cref{fig:regression_real_exp} we show the realized trajectory for the third episode of the racecar. We see that already after three episodes, \alg learns to drive the car close to the target whereas not incorporating the prior leads to exploration away from the target position. Similar behavior is observed on hardware (c.f.,~\Cref{fig:hw_online_rl}), where from episode 1, the policy learned using \alg drives the car close to the target, and at episode 10 it is parking the car nearly perfectly. On the contrary, not incorporating the prior leads to no movement of the car in episode 1 and the car driving away from the target in episode 10. This experiment demonstrates that incorporating a simulation prior in model-based RL leads to directed exploration which results in better sample-efficiency.



\section{Conclusion}
\looseness -1 In this work, we address the sim-to-real gap in robotics by harnessing both domain knowledge and data. Leveraging ideas from Bayesian inference in functional space, we
propose \alg, a simple and tractable algorithm to incorporate physical simulation priors in learning neural network robot dynamics. Our resulting algorithm benefits from both the expressiveness of neural networks and the domain knowledge from physical simulators. We show that \alg results in significantly better sample efficiency for supervised learning of robot dynamics and model-based RL. 

\vspace{1em}
\textbf{Future Work.} \looseness=-1
Here, we use a simple Gaussian approximation for score estimation. There are a range of other score estimation techniques \citep[e.g.][]{shi2018spectral, zhou2020nonparametric}. We leave a detailed comparison among the different methods as future work. Moreover, we regard applying \alg to more high-dimensional systems from robotics or other scientific domains as an interesting avenue for future work.



\section{Acknowledgments}
This research was supported by the European Research Council (ERC) under the European Union's Horizon 2020 research and innovation program grant agreement no.\ 815943, the Swiss National Science Foundation under NCCR Automation, grant agreement 51NF40 180545, and the Microsoft Swiss Joint Research Center. Jonas Rothfuss was supported by an Apple Scholars in AI/ML fellowship.
\bibliography{root}

\begin{thebibliography}{10}
\providecommand{\url}[1]{#1}
\csname url@rmstyle\endcsname
\providecommand{\newblock}{\relax}
\providecommand{\bibinfo}[2]{#2}
\providecommand\BIBentrySTDinterwordspacing{\spaceskip=0pt\relax}
\providecommand\BIBentryALTinterwordstretchfactor{4}
\providecommand\BIBentryALTinterwordspacing{\spaceskip=\fontdimen2\font plus
\BIBentryALTinterwordstretchfactor\fontdimen3\font minus \fontdimen4\font\relax}
\providecommand\BIBforeignlanguage[2]{{%
\expandafter\ifx\csname l@#1\endcsname\relax
\typeout{** WARNING: IEEEtran.bst: No hyphenation pattern has been}%
\typeout{** loaded for the language `#1'. Using the pattern for}%
\typeout{** the default language instead.}%
\else
\language=\csname l@#1\endcsname
\fi
#2}}

\bibitem{widmer2023tuning}
D.~Widmer, D.~Kang, B.~Sukhija, J.~H{\"u}botter, A.~Krause, and S.~Coros, ``Tuning legged locomotion controllers via safe bayesian optimization,'' \emph{CORL}, 2023.

\bibitem{hwangbo}
J.~Hwangbo, J.~Lee, A.~Dosovitskiy, D.~Bellicoso, V.~Tsounis, and M.~H. Vladlen~Koltun, ``Learning agile and dynamic motor skills for legged robots,'' \emph{Science Robotics}, 2019.

\bibitem{singh2022reinforcement}
B.~Singh, R.~Kumar, and V.~P. Singh, ``Reinforcement learning in robotic applications: a comprehensive survey,'' \emph{AI Review}, 2022.

\bibitem{deisenrothPILCOModelBasedDataEfficient2011}
M.~Deisenroth and C.~Rasmussen, ``{{PILCO}}: {{A Model-Based}} and {{Data-Efficient Approach}} to {{Policy Search}}.''

\bibitem{polydoros2017survey}
A.~S. Polydoros and L.~Nalpantidis, ``Survey of model-based reinforcement learning: Applications on robotics,'' \emph{Journal of Intelligent \& Robotic Systems}, 2017.

\bibitem{wu2023daydreamer}
P.~Wu, A.~Escontrela, D.~Hafner, P.~Abbeel, and K.~Goldberg, ``Daydreamer: World models for physical robot learning,'' in \emph{CORL}, 2023.

\bibitem{bhardwaj2023data}
A.~Bhardwaj, J.~Rothfuss, B.~Sukhija, Y.~As, M.~Hutter, S.~Coros, and A.~Krause, ``Data-efficient task generalization via probabilistic model-based meta reinforcement learning,'' \emph{arXiv}, 2023.

\bibitem{wang2019function}
Z.~Wang, T.~Ren, J.~Zhu, and B.~Zhang, ``{Function space particle optimization for Bayesian neural networks},'' in \emph{ICLR}, 2019.

\bibitem{sun2019functional}
S.~Sun, G.~Zhang, J.~Shi, and R.~Grosse, ``Functional variational bayesian neural networks,'' in \emph{ICLR}, 2019.

\bibitem{zhao2020sim}
W.~Zhao, J.~P. Queralta, and T.~Westerlund, ``Sim-to-real transfer in deep reinforcement learning for robotics: a survey,'' in \emph{IEEE SSCI}, 2020.

\bibitem{duan2016rl}
Y.~Duan, J.~Schulman, X.~Chen, P.~L. Bartlett, I.~Sutskever, and P.~Abbeel, ``Rl2: Fast reinforcement learning via slow reinforcement learning,'' in \emph{ICLR}, 2017.

\bibitem{finn2017model}
C.~Finn, P.~Abbeel, and S.~Levine, ``Model-agnostic meta-learning for fast adaptation of deep networks,'' in \emph{ICML}, 2017.

\bibitem{finn2018probabilistic}
C.~Finn, K.~Xu, and S.~Levine, ``Probabilistic model-agnostic meta-learning,'' in \emph{NeurIPS}, 2018.

\bibitem{ljungSystemID}
L.~Ljung, \emph{System Identification}.\hskip 1em plus 0.5em minus 0.4em\relax John Wiley \& Sons, Ltd, 1999.

\bibitem{li2013terradynamics}
C.~Li, T.~Zhang, and D.~I. Goldman, ``A terradynamics of legged locomotion on granular media,'' \emph{science}, 2013.

\bibitem{moeckel2013gait}
R.~Moeckel, Y.~N. Perov, A.~T. Nguyen, M.~Vespignani, S.~Bonardi, S.~Pouya, A.~Sproewitz, J.~van~den Kieboom, F.~Wilhelm, and A.~J. Ijspeert, ``Gait optimization for roombots modular robots—matching simulation and reality,'' in \emph{IROS}, 2013, pp. 3265--3272.

\bibitem{tan2016simulation}
J.~Tan, Z.~Xie, B.~Boots, and C.~K. Liu, ``Simulation-based design of dynamic controllers for humanoid balancing,'' in \emph{IROS}.\hskip 1em plus 0.5em minus 0.4em\relax IEEE, 2016, pp. 2729--2736.

\bibitem{zhu2017model}
S.~Zhu, A.~Kimmel, K.~E. Bekris, and A.~Boularias, ``Model identification via physics engines for improved policy search,'' \emph{IJCAI}, 2017.

\bibitem{gofetch}
S.~Zimmermann, R.~Poranne, and S.~Coros, ``Go fetch! - dynamic grasps using boston dynamics spot with external robotic arm,'' in \emph{ICRA}, 2021.

\bibitem{pastor2013learning}
P.~Pastor, M.~Kalakrishnan, J.~Binney, J.~Kelly, L.~Righetti, G.~Sukhatme, and S.~Schaal, ``Learning task error models for manipulation,'' in \emph{ICRA}, 2013.

\bibitem{ha2015reducing}
S.~Ha and K.~Yamane, ``Reducing hardware experiments for model learning and policy optimization,'' in \emph{ICRA}, 2015.

\bibitem{HUANG20178969}
H.~Huang, W.~He, and S.~Zhang, ``Neural network control of a robotic manipulator with time-varying output constraints by state feedback,'' \emph{IFAC}, 2017.

\bibitem{schperberg2023real}
A.~Schperberg, Y.~Tanaka, F.~Xu, M.~Menner, and D.~Hong, ``Real-to-sim: Predicting residual errors of robotic systems with sparse data using a learning-based unscented kalman filter,'' in \emph{UR}, 2023.

\bibitem{narendraNNSysID}
K.~Narendra and K.~Parthasarathy, ``Identification and control of dynamical systems using neural networks,'' \emph{IEEE Transactions on Neural Networks}, 1990.

\bibitem{SJOBERG1994359}
J.~Sjöberg, H.~Hjalmarsson, and L.~Ljung, ``Neural networks in system identification,'' \emph{IFAC Proceedings Volumes}, 1994.

\bibitem{nagabandi2018}
A.~Nagabandi, G.~Kahn, R.~S. Fearing, and S.~Levine, ``Neural network dynamics for model-based deep reinforcement learning with model-free fine-tuning,'' in \emph{ICRA}, 2018.

\bibitem{bernSoftRobotControl2020}
J.~M. Bern, Y.~Schnider, P.~Banzet, N.~Kumar, and S.~Coros, ``Soft {{Robot Control With}} a {{Learned Differentiable Model}},'' in \emph{{RoboSoft}}, 2020.

\bibitem{sukhija2023gradient}
B.~Sukhija, N.~K{\"o}hler, M.~Zamora, S.~Zimmermann, S.~Curi, A.~Krause, and S.~Coros, ``Gradient-based trajectory optimization with learned dynamics,'' in \emph{ICRA}, 2023.

\bibitem{chua2018deep}
K.~Chua, R.~Calandra, R.~McAllister, and S.~Levine, ``{Deep Reinforcement Learning in a Handful of Trials using Probabilistic Dynamics Models},'' in \emph{NeurIPS}, 2018.

\bibitem{curi2020efficient}
S.~Curi, F.~Berkenkamp, and A.~Krause, ``Efficient model-based reinforcement learning through optimistic policy search and planning,'' \emph{NeurIPS}, 2020.

\bibitem{sekar2020planning}
R.~Sekar, O.~Rybkin, K.~Daniilidis, P.~Abbeel, D.~Hafner, and D.~Pathak, ``Planning to explore via self-supervised world models,'' in \emph{ICLR}, 2020.

\bibitem{treven2024ocorl}
L.~Treven, J.~Hübotter, B.~Sukhija, F.~Dörfler, and A.~Krause, ``Efficient exploration in continuous-time model-based reinforcement learning,'' \emph{NeurIPS}, 2024.

\bibitem{sukhija2024optimistic}
B.~Sukhija, L.~Treven, C.~Sancaktar, S.~Blaes, S.~Coros, and A.~Krause, ``Optimistic active exploration of dynamical systems,'' \emph{NeurIPS}, 2024.

\bibitem{rothfuss2023hallucinated}
J.~Rothfuss, B.~Sukhija, T.~Birchler, P.~Kassraie, and A.~Krause, ``Hallucinated adversarial control for conservative offline policy evaluation,'' \emph{UAI}, 2023.

\bibitem{mackay1992bayesian}
D.~J. MacKay, ``Bayesian methods for adaptive models,'' Ph.D. dissertation, California Institute of Technology, 1992.

\bibitem{neal2012bayesian}
R.~M. Neal, \emph{Bayesian learning for neural networks}.\hskip 1em plus 0.5em minus 0.4em\relax Springer Science \& Business Media, 2012.

\bibitem{graves2011practical}
A.~Graves, ``Practical variational inference for neural networks,'' \emph{NeurIPS}, 2011.

\bibitem{blundell2015weight}
C.~Blundell, J.~Cornebise, K.~Kavukcuoglu, and D.~Wierstra, ``Weight uncertainty in neural network,'' in \emph{ICML}, 2015.

\bibitem{gal2016dropout}
Y.~Gal and Z.~Ghahramani, ``Dropout as a bayesian approximation: Representing model uncertainty in deep learning,'' in \emph{ICML}, 2016.

\bibitem{zellers2018swag}
R.~Zellers, Y.~Bisk, R.~Schwartz, and Y.~Choi, ``Swag: A large-scale adversarial dataset for grounded commonsense inference,'' \emph{arXiv}, 2018.

\bibitem{goan2020bayesian}
E.~Goan and C.~Fookes, ``Bayesian neural networks: An introduction and survey,'' \emph{Case Studies in Applied Bayesian Data Science}, 2020.

\bibitem{lakshminarayanan2017simple}
B.~Lakshminarayanan, A.~Pritzel, and C.~Blundell, ``Simple and scalable predictive uncertainty estimation using deep ensembles,'' \emph{NeurIPS}, 2017.

\bibitem{Liu2016}
Q.~Liu and D.~Wang, ``{Stein Variational Gradient Descent: A General Purpose Bayesian Inference Algorithm},'' in \emph{NeurIPS}, 2016.

\bibitem{d2021repulsive}
F.~D'Angelo and V.~Fortuin, ``Repulsive deep ensembles are bayesian,'' \emph{NeurIPS}, 2021.

\bibitem{Blei2016}
D.~M. Blei, A.~Kucukelbir, and J.~D. McAuliffe, ``Variational inference: {A} review for statisticians,'' \emph{Journal of the American Statistical Association}, 2017.

\bibitem{chen2018unified}
C.~Chen, R.~Zhang, W.~Wang, B.~Li, and L.~Chen, ``A unified particle-optimization framework for scalable bayesian sampling,'' 2018.

\bibitem{rasmussen2003gaussian}
C.~E. Rasmussen and C.~K.~I. Williams, \emph{{Gaussian processes in machine learning}}.\hskip 1em plus 0.5em minus 0.4em\relax MIT Press, 2006.

\bibitem{stochastic_differential_equations}
B.~Øksendal, \emph{{Stochastic Differential Equations: An Introduction with Applications}}, 2000.

\bibitem{kabzan2020amz}
J.~Kabzan, M.~I. Valls, V.~J. Reijgwart, H.~F. Hendrikx, C.~Ehmke, M.~Prajapat, A.~B{\"u}hler, N.~Gosala, M.~Gupta, R.~Sivanesan, \emph{et~al.}, ``Amz driverless: The full autonomous racing system,'' \emph{Journal of Field Robotics}, 2020.

\bibitem{pacejka1992magic}
H.~B. Pacejka and E.~Bakker, ``The magic formula tyre model,'' \emph{Vehicle system dynamics}, pp. 1--18, 1992.

\bibitem{sac}
T.~Haarnoja, A.~Zhou, P.~Abbeel, and S.~Levine, ``Soft actor-critic: Off-policy maximum entropy deep reinforcement learning with a stochastic actor,'' in \emph{ICML}, 2018.

\bibitem{janner2019trust}
M.~Janner, J.~Fu, M.~Zhang, and S.~Levine, ``When to trust your model: Model-based policy optimization,'' \emph{NeurIPS}, 2019.

\bibitem{shi2018spectral}
J.~Shi, S.~Sun, and J.~Zhu, ``A spectral approach to gradient estimation for implicit distributions,'' in \emph{ICML}, 2018.

\bibitem{zhou2020nonparametric}
Y.~Zhou, J.~Shi, and J.~Zhu, ``Nonparametric score estimators,'' in \emph{ICML}, 2020.

\end{thebibliography}

\end{document}